\documentclass[9.5pt,journal,finalsubmission,final,twocolumn,twoside]{IEEEtran}

\usepackage{cite}
\usepackage{graphicx}
\usepackage{float}
\usepackage{math}
\usepackage{amsmath}
\usepackage{amsfonts}
\usepackage{amssymb}
\usepackage{cleveref}
\usepackage{url}
\usepackage{algorithm}
\usepackage{algpseudocode}

\usepackage[final]{review}
\setcoverletter{AnswerReviewer2.tex}
\setrevision{2}

\begin{document}
\title{Robust Head-Pose Estimation Based on Partially-Latent Mixture of Linear Regressions}

\author{\IEEEauthorblockN{Vincent Drouard\IEEEauthorrefmark{1},
Radu Horaud\IEEEauthorrefmark{1}},
Antoine Deleforge\IEEEauthorrefmark{2},
Sil\`{e}ye Ba\IEEEauthorrefmark{1}\IEEEauthorrefmark{3} and
Georgios Evangelidis\IEEEauthorrefmark{1}\IEEEauthorrefmark{4}\\
\IEEEauthorblockA{\IEEEauthorrefmark{1}INRIA Grenoble Rh\^{o}ne-Alpes, Montbonnot Saint-Martin, France\\
\IEEEauthorrefmark{2}INRIA Rennes Bretagne Atlantique, Rennes, France\\
\IEEEauthorrefmark{3}VideoStitch, Paris, France 
\IEEEauthorrefmark{4}DAQRI Int, Dublin, Ireland}
% <-this % stops an unwanted space
\thanks{This work has been funded by the European Research Council through the ERC Advanced Grant VHIA \#340113. R. Horaud acknowledges support from a XEROX University Affairs Committee (UAC) grant (2015-2017).}}

%\markboth{IEEE Transactions on Image Processing,~Vol.~26, No.~X, 2017}% {Drouard \MakeLowercase{\textit{et al.}}: Robust Head-Pose Estimation Based on Partially-Latent Mixture of Linear Regressions}

% \IEEEpubid{0000--0000/00\$00.00~\copyright~2015 IEEE}


% for Transactions on Magnetics papers, we must declare the abstract and
% index terms PRIOR to the title within the \IEEEtitleabstractindextext
% IEEEtran command as these need to go into the title area created by
\IEEEtitleabstractindextext{
\begin{abstract}
	Head-pose estimation has many applications, such as social event analysis, human-robot and human-computer interaction, driving assistance, and so forth. Head-pose estimation is challenging because it must cope with changing illumination conditions, variabilities in face orientation and in appearance, partial occlusions of facial landmarks, as well as bounding-box-to-face alignment errors. We propose tu use a mixture of linear regressions with partially-latent output. This regression method learns to map high-dimensional feature vectors (extracted from bounding boxes of faces) onto the joint space of head-pose angles and bounding-box shifts, such that they are robustly predicted in the presence of unobservable phenomena. We describe in detail the mapping method that combines the merits of unsupervised manifold learning techniques and of mixtures of regressions. We validate our method with three publicly available datasets and we thoroughly benchmark four variants of the proposed algorithm with several state-of-the-art head-pose estimation methods.
\end{abstract}

\begin{IEEEkeywords}
	Head pose, face detection, mixture of linear regressions, manifold learning, expectation-maximization.
\end{IEEEkeywords}}

\maketitle

 \IEEEdisplaynontitleabstractindextext

% \fi
\IEEEpeerreviewmaketitle

	\section{Introduction}
	\label{section:introduction}	
	
%\IEEEPARstart{H}ead
Head pose is an important visual cue in many scenarios such as social-event analysis~\cite{sabanovic06robots}, human-robot interaction (HRI)~\cite{goodrich07human} or driver-assistance systems~\cite{murphy2007head} to name a few. For example, in social-event analysis, 3D head pose information drastically helps to determine the interaction between people and to extract the visual focus of attention~\cite{murphy2009head}. The pose is typically expressed by three angles (pitch, yaw, roll) that describe the \addnote[label-intro-2]{1}{orientation with respect to a head-centered frame}. 
%The pitch and yaw represent the rotation around the horizontal and vertical axes of the head respectively, whereas the roll angle represents the in-plane rotation of the face.
The estimation of the pose parameters is challenging for many reasons. \addnote[label-intro-3]{1}{Algorithms for }head-pose estimation must be invariant to changing illumination conditions, to the background scene, to partial occlusions, and to inter-person and intra-person variabilities. In most application scenarios, faces have small support area, \ie bounding boxes, typically of the order of $100\times 100$ pixels. Even if the face bounding box is properly detected, one has to extract the pose angles from low-resolution data. 

Recent advances in computer vision have shown the relevance of representing an image patch with a feature vector, \eg SIFT \cite{lowe2004distinctive}, HOG \cite{dalal2005histograms}, SURF \cite{bay2008speeded}, or one of their variants. The rationale of representing faces in a high-dimensional feature space is that the latter supposedly embeds a low-dimensional manifold parameterized by the pose parameters, or the head-pose manifold, \eg \cite{osadchy2007synergistic,foytik2013atwo,DeleforgeForbesHoraud2015}. Hence, several attempts were carried out in order to cast the problem at hand into various frameworks, such as manifold learning (unsupervised) \cite{tenenbaum2000global}, regression \cite{DeleforgeForbesHoraud2015,drouard2015head}, convolutional neural networks \cite{osadchy2007synergistic} (supervised), or dimensionality reduction followed by regression \cite{foytik2013atwo}, to cite just a few.

While the papers just cited yield interesting and promising results, there are several major issues associated with representing faces with high-dimensional feature vectors, issues that have not been properly addressed. Indeed and as already mentioned, these vectors contain many underlying phenomena other than pose, \eg illumination, appearance, shape, background, clutter, etc. Hence, one major challenge is to be able to remove these other pieces of information and to retain only head-pose information. Another drawback is that head pose relies on face detection, a process that amounts to finding a bounding box that contains the face and which is invariant to face orientation.

Take for example the case of finding pose parameters using linear or non-linear manifold learning followed by regression. 
This is usually justified by the fact that high-dimensional-to-low-dimensional (high-to-low) regression has to estimate a very large number of parameters, typically of the order of $D^2$ where $D$ is the dimension of the feature space. This in turn requires a huge training dataset. 
Moreover, this sequential way of doing presents the risk to map the input onto an intermediate low-dimensional space that does not necessarily contain the information needed for the finally desired output -- head pose. Finally, the estimation of the pose angles in two distinct steps cannot be conveniently expressed in a single optimization formulation.

\begin{figure*}[t!]
	\includegraphics[width=\textwidth]{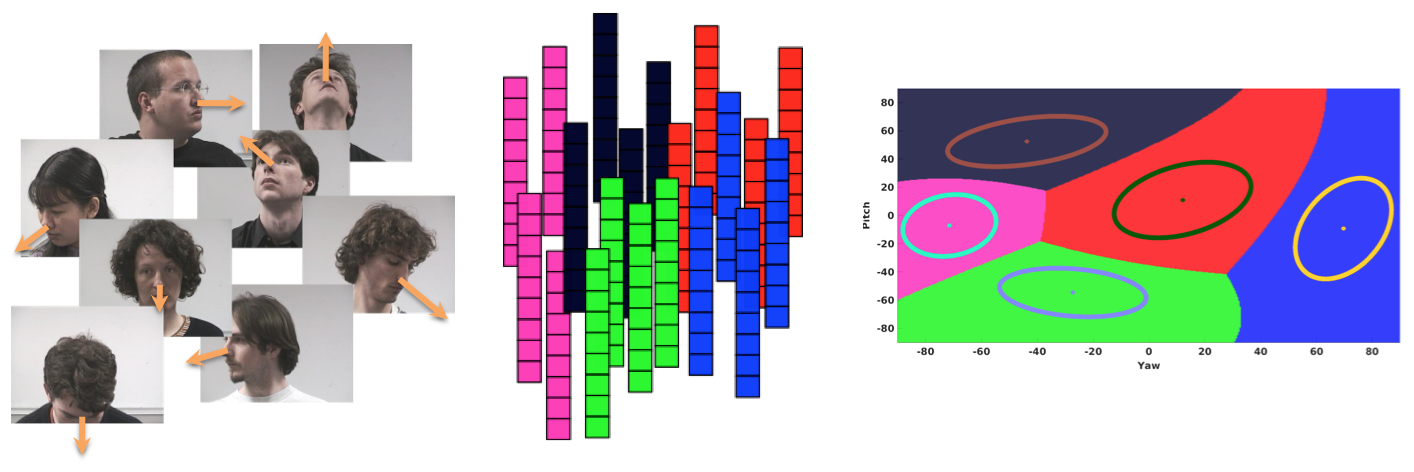}\\
	 \includegraphics[width=\textwidth]{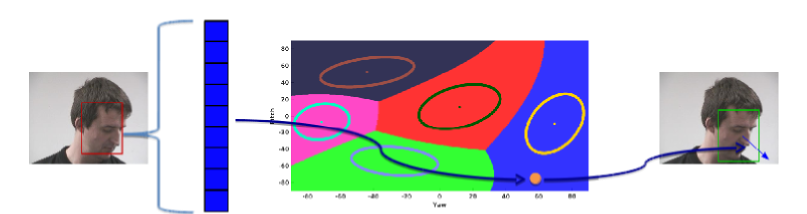}
	\caption{Pipeline of the proposed supervised head-pose estimation method. \textbf{Top:} the parameters of a mixture of linear regressions are learnt from faces annotated with their poses (left). The result of this learning is a simultaneous partitioning of both the high-dimensional input (high-dimensional feature vectors shown in the middle) and low-dimensional output (two-dimensional parameter space shown on the right), such that each region in this partition corresponds to an affine mapping between the input and the output. Moreover, the output is modeled by a Gaussian mixture and each region corresponds to a mixture component. This yields a predictive distribution that can then be used to predict an output from a test input. \textbf{Bottom:} A face detector is used to localize a bounding box (left, shown in red) from which a HOG descriptor, namely a high-dimensional feature vector, is extracted. Using the predictive distribution just mentioned, it is then possible to estimate the head-pose parameters (yaw and pitch in this example). Additionally, it is also possible to refine the bounding-box location such that the latter is optimally aligned with the face (right, shown in green).}
	\label{figure:pipeline}
\end{figure*}

Supervised pose estimation also requires an annotated dataset of faces with their bounding-box locations and the corresponding pose parameters. 
The large majority of existing methods relies on manually annotated faces for training. In particular this ensures good bounding-box-to-face alignment, \ie the face center coincides with the bounding-box center and the bounding box does not contain too many background pixels. This alignment requirement has, however, an important drawback, namely that bounding boxes found by face detection techniques do not correspond to the annotated ones. This means that feature vectors obtained with a face detector do not necessarily belong to the space of feature vectors obtained by manual annotation and used for training. Hence, there is a risk that the head pose, predicted from feature vectors associated with automatic face detection, is incorrectly estimated.

 \addnote[label-intro-4]{1}
{In this paper we propose to learn with both head-pose parameters and bounding-box-to-face alignments, such that, at runtime both the head-pose angles and bounding-box shifts are predicted. This ensures that the bounding-box-to-face alignments predicted with our method are similar with those used for training. Consequently, background variations have minimal influence on the observed feature vector from which the pose is being predicted. This prevents from pose-estimation errors due to discrepancies that occur between manually annotated faces (used for training) and automatically detected faces (used for prediction)}. 
We adopt a high-dimensional description of faces based on HOG features,
hence we need to solve a high-to-low regression problem. Rather then performing dimensionality reduction and regression in two distinct steps, we propose to use a generative model that unifies these two steps in a single one. 

More precisely, we propose to adopt the partially-latent mixture of linear regression model of \cite{DeleforgeForbesHoraud2015}. When applied to head pose estimation from feature vectors, this method has two advantages: (i)~it solves high-dimensional to low-dimensional regression in a single process, without the need of a dimensionality reduction pre-processing step, and (ii)~the method incorporates \textit{latent variable augmentation}. More precisely, the output variables are only partially observed such that the regression function can be trained with partially observed outputs. The method adopts an inverse regression strategy  \cite{li1991sliced}: the roles of the input and output variables are interchanged, such that high-to-low regression is replaced with low-to-high regression, followed by high-to-low prediction.
As illustrated by our experiments, \eg Fig.~\ref{figure:pipeline}, we found that the inclusion of a partially-latent output is particularly relevant whenever the high-dimensional input is corrupted by irrelevant information that cannot be easily annotated, \eg facial expressions, changes in illumination conditions, changes in background, etc.

Another contribution is a thorough experimental evaluation and validation of  the merits of \cite{DeleforgeForbesHoraud2015} when the task is to recover low-dimensional parameterizations of manifolds embedded in high-dimensional feature vector representations of image patches. We note that this task is ubiquitous in computer vision and image processing, hence a  method that unifies dimensionality reduction and regression in a principled manner is of primary importance and of great utility for practitioners.
Moreover, with respect to \cite{DeleforgeForbesHoraud2015} we provide more detailed mathematical derivations as well as a unified view of the algorithms used for training, such that the method can be readily reproduced by others.\footnote{Supplementary material, that includes a complete Matlab package and examples of trained regression functions, is publicly available at \url{https://team.inria.fr/perception/research/head-pose}.}

The remainder of the paper is organized as follows. Section~\ref{section:related} discusses related work. The regression method is described in Section~\ref{section:probabilistic} and the algorithmic details can be found in Appendix~\ref{appendix:em-gllim}. Section~\ref{section:implementation} describes implementation details. Experimental results are presented and discussed in Section~\ref{section:experiments}. Finally, Section~\ref{section:conclusion} draws some conclusions for future work.

	\section{Related Work}
	\label{section:related}
	% Related work section

Head-pose estimation has been very well investigated for the past decades and several powerful methods have been developed. 
The head-pose estimation literature was surveyed a few years ago \cite{murphy2009head}, however this survey does not include methods based on depth data as these papers were published after 2009. For convenience we grouped the head-pose methods into four categories: (i)~methods that are based on depth images, (ii)~methods based on manifold learning, (iii)~methods based on regression, and (iv) methods that combine pose with face detection and localization.

The recent advent of depth cameras enabled fast development of depth-based head-pose methods.
Depth data allow to overcome some of the drawbacks of RGB data, such as illumination problems and facial landmarks detection, which is more reliable.  One of the first methods that uses depth data is \cite{seemann2004head}, were a depth map of the head region is combined with a color histogram and used to train a neural network.
Random forest regression is proposed in \cite{fanelli2013random} to estimate both head pose and facial landmarks. Random forest regression is also used in \cite{Peng201592} where RGB SIFT descriptors are combined with 3D HOG descriptors. \addnote[label-relat-2]{1}{More recently \cite{ghiass2015highly} released a new method combining RGB and depth images to infer head pose. The RGB image is used to find facial landmarks, the 3D positions of the landmarks are used to fit a plane on the face that will be used to extract the 3D points that belong to the face. Using the face 3D point clouds, a morphable model of a face is map on it using optimization methods to estimate head orientation parameters. }A general observation about these methods is that depth information is merely used to disambiguate photometric data and that depth data cannot be used alone for head pose.

Several authors proposed to use manifold learning, namely finding a low-dimensional output space of head poses from a high-dimensional input space of feature vectors. Nevertheless, the output variables do not necessarily correspond to the pose angles, so one has to farther learn in a supervised way the mapping between the manifold-learning output and the desired space spanned by the pose parameters. This has been achieved in various ways, \eg \cite{srinivasan2002head,raytchev2004head,hu2005head,li2007query,benabdelkader2010robust,foytik2013atwo,sundara2015head}. As already mentioned, these two step methods suffer from the fact that unsupervised manifold-learning techniques do not guarantee that the predicted output space contains the information needed for head pose.

Among the regression methods used for head pose are Gaussian process regression (GPR) \cite{marin2014detecting}, support vector regression (SVR) \cite{murphy2007head}, partial least squares (PLS) \cite{sharma2011bypassing} and and kernel PLS \cite{haj2012partial}. Both  \cite{marin2014detecting} and  \cite{murphy2007head} estimate the pose angles independently, so several regression functions must be learned, one for each angle, hence correlations between these parameters cannot be taken into account. Another drawback of all kernel methods is that they require the design of a kernel function with its hyper-parameters, which must be either manually selected or properly estimated using non-convex optimization techniques. 

PLS and kernel PLS proceed in two steps. First, both the input and the output are projected onto low-dimensional latent subspaces by maximizing the covariance between the projected input and the projected output. Second, a linear regression between these two latent subspaces is estimated. The performance of PLS is subject to the relationship between the covariance matrices of input and output variables and to the eigen structure of the covariance of the input variable \cite{NaikTsai2000}. The advantage of the proposed method is that it estimates a mixture of linear regressions directly from the input and output variables. 

The methods described so far use manually annotated images for training and face detectors for testing. As already discussed, this could lead to pose estimation errors because a test feature vector may not lie in the subspace spanned by the feature vectors used for training. One way to deal with this problem is to combine face detection and pose estimation in a single task. For example, \cite{zhu2012face} considers face templates for all possible poses which are then fed into a cascaded classifier.

Convolutional neural network (CNN) architectures were also proposed in the recent past \cite{osadchy2007synergistic}, \cite{ahn2014real}. \cite{osadchy2007synergistic} considers a fixed image sub-window at all locations and scales. The network consists of $64,000$ weights and kernel coefficients that need to be estimated, and both face and non-face samples must be considered. Altogether, training the network with $52,000$ positives and $52,000$ negatives involves non-linear optimization and takes 26 hours on a 2GHz Pentium 4. \addnote[label-relat-1]{1}{\cite{ahn2014real} proposed a CNN architecture composed of four convolutional layers with max-pooling on the first two layers; the activation function is the hyperbolic tangent which yields good convergence during the training phase. Small input RGB images ($32\times32$ pixels) and small filters ($5 \times 5$ pixels) are used in order to overcome the limitation of the training dataset. The network is trained using $13,500$ face patches extracted from the dataset}. \addnote[label-relat-4]{1}{More recently, \cite{Liu20163d} proposed to simulate a dataset of head poses in order to train a CNN. Then they use the trained network to estimate head pose from real color images using the BIWI dataset \cite{fanelli2013random}. They show that when trained using the BIWI dataset the CNN approach yields results similar to \cite{drouard2015head} and that the accuracy is improved, by a factor of 2, when a large set of simulated images are used for training.
}

%\cite{murphy2010} try to deal with this issue by using a face descriptor that is robust to face misalignment, they use Localized Orientations Gradient (LOG), they don't require any additional training to take into account the error in the face localization. 

The problem of bounding-box-to-face miss-alignment  was discussed and addressed in \cite{haj2012partial}. First, kernel PLS is trained on manually extracted bounding boxes and associated pose parameters, and second, head-pose prediction is applied to the bounding box found by a face detector, as well as to a number of shifted bounding boxes. The bounding box that produces the minimum PLS residual is selected for predicting the pose parameters using the trained regression.
This results in a time-consuming estimator since both many bounding boxes must be considered and latent-space PLS projections must be performed at runtime. The advantage of our method with respect to \cite{haj2012partial} is that bounding box shifting is embedded in the learning process such that the optimal shift becomes part of the output, which yields a computationally efficient head pose predictor.

\addnote[label-relat-3]{1}{A short version of this paper was recently published \cite{drouard2015head}. The main addition with respect to \cite{drouard2015head} is the use of a partially-latent response variable associated with the proposed inverse regression method. This can be viewed as an augmented model and its incorporation in our formulation yields excellent results because it enables robustness to phenomena that make the head-pose estimation problem difficult. Both the proposed method (Section~\ref{section:probabilistic}) and the algorithm (Appendix~\ref{appendix:em-gllim}) are described in more detail than in  \cite{drouard2015head}, which makes the paper self-contained, self-explanatory, and enables others to easily reproduce the results. Additionally, Section~\ref{section:experiments} contains an extended set of results and comparisons, using three publicly available datasets. }

	\section{Partially-Latent Mixture of Linear Regressions}
	\label{section:probabilistic}
	In this section we summarize the generative \textit{inverse} regression method proposed in~\cite{DeleforgeForbesHoraud2015}. The method is referred to as Gaussian locally linear mapping (GLLiM). GLLiM interchanges the roles of the input and output variables, such that a \textit{low-dimensional to high-dimensional} regression problem is solved instead of a high-dimensional to low-dimensional one. The immediate consequence of using such an inverse regression model is a dramatic reduction in the number of model parameters, thus facilitating the task of training. 

\addnote[label-prob-1]{2}{
An additional advantage of this method is that it can be trained by adding a latent part to the output: while the high-dimensional input remains fully observed, the low-dimensional output is a concatenation of a multivariate observed variable and a multivariate latent variable. This is referred to as hybrid-GLLiM (or hGLLiM). The latent part of the output variable has a principled mathematical definition (see below) but it does not necessarily have a physical meaning. The main idea (of introducing a partially latent output) relies on the fact that variabilities present in the high-dimensional face descriptor depend on head pose, on face to bounding-box alignment, and on other phenomena (face shapes, facial expressions, gender, hair and skin color, age, etc.) that are not relevant for the task at hand. The latent part of the regression output has the interesting feature of gathering information other than pose parameters and bounding-box parameters.
}

\subsection{Inverse Regression}
\label{subsec:invreg_nonhybrid}
Let $\Xvect$, $\Yvect$ be two random variables, such that $\Xvect\in\mathbb{R}^L$ denotes the low-dimensional output, \eg pose parameters and $\Yvect\in\mathbb{R}^D$ ($D\gg L$) denotes the high-dimensional input, \eg feature vector. The goal is to predict a {\it response} $\Xvect$ given both an {\it input} $\Yvect$ and the model parameters $\thetavect$, \ie $p(\Xvect | \Yvect ; \thetavect)$. 
We consider the \textit{inverse low-to-high} mapping from the output variable $\Xvect$ to the input variable $\Yvect$. This, possibly non-linear, mapping is modeled by a mixture of locally-affine transformations:
\begin{equation}
\label{eq:model_yxz}
  \Yvect=\sum_{k=1}^K \mathbb{I}(Z=k) (\Amat_k \Xvect + \bvect_k +
  \evect_{k}),
\end{equation}
where $\mathbb{I}$ is the indicator function, \addnote[label-prob-2]{1}{and $Z$ is the standard discrete latent variable}: $Z=k$ if and only if $\Yvect$ is the image of
$\Xvect$ by the affine transformation $\Amat_k \Xvect + \bvect_k$, with
$\Amat_k\in\mathbb{R}^{D\times L}$ and $\bvect_k\in\mathbb{R}^D$, and
$\evect_{k}\in\mathbb{R}^D$ is an error vector capturing both the
high-dimensional observation noise and the reconstruction error
due to the piecewise approximation. The missing-data variable $Z$ allows one to write the joint probability of $\Xvect$ and $\Yvect$ as the following mixture:
\begin{align}
 p  (\Yvect=\yvect,\Xvect & =\xvect;\thetavect) = \sum_{k=1}^K p(\Yvect=\yvect | \Xvect=\xvect,Z=k;\thetavect)
\nonumber \\
\label{eq:joint-prob-distr}
& \times p(\Xvect=\xvect | Z=k;\thetavect) p(Z=k;\thetavect),
\end{align}
where $\thetavect$ denotes the model parameters and $\yvect$ and $\xvect$ denote realizations of $\Yvect$ and $\Xvect$ respectively. Assuming that $\evect_k$ is a zero-mean Gaussian variable with diagonal covariance matrix $\Sigmamat_k\in\mathbb{R}^{D\times D}$ with diagonal entries $\sigma_{k1}, \dots, \sigma_{kD}$, we obtain that
%The conditional probability $p(\yvect|\xvect ; \thetavect)$ in (\ref{eq:joint-prob-distr}) can express (\ref{eq:model_yxz}) in a probabilistic way provided that $\evect_{k}$ is a zero-mean Gaussian variable, namely
%\begin{equation}
%\label{eq:model_pYxz}
$  p(\yvect|\xvect,Z=k;\thetavect) =
 \mathcal{N}(\yvect ;\Amat_k \xvect + \bvect_k ,\Sigmamat_k)$.
%\end{equation}
%where $\Sigmamat_k\in\mathbb{R}^{D\times D}$ is a covariance matrix.
If we further assume that $\Xvect$ follows a mixture of Gaussians via the same assignment $Z=k$, we can write that
%\begin{align}
$ p(\xvect |Z=k; \thetavect)  = \mathcal{N}(\xvect ; \cvect_k,\Gammamat_k)$ and
$ p(Z=k; \thetavect) = \pi_k \label{eq:model_pXZ}$,
%\end{align}
where $\cvect_k\in\mathbb{R}^L$, $\Gammamat_k\in\mathbb{R}^{L\times
L}$ and $\sum_{k=1}^K \pi_k=1$. Note that this representation induces a partition of $\mathbb{R}^{L}$ into $K$ regions ${\cal R}_k$, where ${\cal R}_k$ is the region where the transformation $(\Amat_k, \bvect_k)$ is most likely invoked, \eg Fig.~\ref{figure:pipeline}. This model is described by the parameter set
\begin{equation}
\label{eq:theta-def}
\thetavect = \{\cvect_k,\Gammamat_k,\pi_k,\Amat_k, \bvect_k,\Sigmamat_k\}_{k=1}^K.
\end{equation} 

\addnote[inverse-def]{2}{
Notice that the number of model parameters $\thetavect$ is dictated by the number of parameters of a multivariate Gaussian distribution and by the number of Gaussian components ($K$). However, the number of parameters of an unconstrained GMM is quadratic in the dimension of the variable. Hence, the size of the training dataset required to reliably estimate a conventional GMM would become prohibitively high for the dimension considered in the paper ($D=1888$). This is why an inverse regression strategy is adopted, making the number of parameters linear in the input variable dimension rather than quadratic. This drastically reduces the model size in practice, making it tractable. 
}

\subsection{Inverse Regression with Partially Latent Output}
We now extend the previous model such that one can train the inverse regression in the presence of partially latent output: hybrid-GLLiM. While the high-dimensional variable $\Yvect$ remains unchanged, \ie fully observed, the low-dimensional variable is a concatenation of an observed variable $\Tvect\in\mathbb{R}^{L_t}$ and a latent variable $\Wvect\in\mathbb{R}^{L_w}$, namely $\Xvect=[\Tvect ; \Wvect]$, where $[\cdot ; \cdot]$ denotes vertical vector concatenation and with $L_t+L_w=L$. Hybrid-GLLiM can be seen as a latent-variable augmentation of standard regression. It can also be seen as a semi-supervised dimensionality reduction method since the unobserved low-dimensional variable $\Wvect$ must be recovered from realizations of the observed variables $\Yvect$ and $\Tvect$.

The decomposition of $\Xvect$ implies that some of the model parameters must be decomposed as well, namely $\cvect_k$, $\Gammamat_k$ and $\Amat_k$. Assuming the independence of $\Tvect$ and $\Wvect$ given $Z$ we have
\begin{equation}
	\cvect_k = \begin{pmatrix} \cvect_k^t \\ \cvect_k^w \end{pmatrix}, \quad \Gammamat_k = \begin{pmatrix} \Gammamat_k^t & 0\\ 0 & \Gammamat_k^w \end{pmatrix}, \quad \Amat_k = \begin{pmatrix}\Amat_k^t & \Amat_k^w \end{pmatrix}.
\end{equation}
It follows that \eqref{eq:model_yxz} rewrites as
\begin{equation}
	\Yvect = \sum_{k=1}^K \mathbb{I}(Z=k) (\Amat_k^t \Tvect + \Amat_k^w \Wvect + \bvect_k 
  + \evect_{k}),
\end{equation}
While the parameters to be estimated are the same, \ie \eqref{eq:theta-def}, there are two sets of missing variables, $Z_{1:N}=\{Z_n\}_{n=1}^{N}\in \{1\dots K\}$ and $\Wvect_{1:N}=\{\Wvect_n \}_{n=1}^{N}\in \mathbb{R}^{L_w}$, associated with the training data $(\yvect,\tvect)_{1:N}=\{\yvect_n,\tvect_n \}_{n=1}^{N}$ given the number of $K$ of affine transformations and the latent dimension $L_w$. Also notice that the means $\{ \cvect_k^w \}_{k=1}^{K}$ and covariances $\{\Gammamat_k^w \}_{k=1}^{K}$ must be fixed to avoid non-identifiability issues. 
Indeed, changing their values corresponds to shifting and scaling the latent variables $\{\Wvect_n \}_{n=1}^{N}$ which are compensated by changes in the parameters of the affine transformations $\{ \Amat_k^w \}_{k=1}^{K}$ and $\{ \bvect_k^w \}_{k=1}^{K}$. This identifiability problem is the same as the one encountered in latent variable models for dimension reduction and is always solved by fixing these parameters. Following \cite{GhahramaniHinton1996} and \cite{TippingBishop1999}, the means and covariances are fixed to zero and to the identity matrix respectively: $\cvect_k^w=\zerovect, \Gammamat_k^w=\Imat, \forall k\in\{1\dots K\}$.

The corresponding EM algorithm consists of estimating the parameter set $\thetavect$ that maximizes
\begin{align}
\thetavect^{(i)} = \argmax\limits_{\thetavect} 
\big(  
\mathbb{E}_Z & [\log p ((\xvect,\yvect, \Wvect, Z)_{1:N}; \thetavect | \nonumber \\
\label{eq:hybrid-gllim-parameters}
& (\xvect, \yvect)_{1:N}; \thetavect^{(i-1)}) ]    
\big).
\end{align}
%which is a generalization of \eqref{eq:gllim-parameters}. 
Using that $\Wvect_{1:N}$ and $\Tvect_{1:N}$ are independent
conditionally on $Z_{1:N}$ and that
$\{\cvect_k^{\textrm{w}}\}_{k=1}^K$ and
$\{\Gammamat_k^{\textrm{w}}\}_{k=1}^K$ are fixed, maximizing (\ref{eq:hybrid-gllim-parameters})
is then equivalent to 
\begin{align} 
\thetavect^{(i)} = \argmax\limits_{\thetavect} \big\{  
\mathbb{E}_{r_{Z}^{(i)}} & [\mathbb{E}_{r_{W|Z}^{(i)}}[\log
p(\yvect_{1:N}\; | \; (\tvect,\Wmat,Z)_{1:N};\thetavect)] \nonumber \\
 & + \log 
p((\tvect,Z)_{1:N};\thetavect)] \big\}, \label{eq:Q-PSM-EM}
\end{align}
where $r_{Z}^{(i)}$ and $r_{W|Z}^{(i)}$ denote
the posterior distributions
\begin{align}
\label{eq:r_wz}
r_{W|Z}^{(i)} &= p(\Wmat_{1:N} |(\yvect,\tvect,Z)_{1:N};\thetavect^{(i-1)}),\\
\label{eq:r_z}
r_{Z}^{(i)} &= p(\Zmat_{1:N} | (\yvect,\tvect)_{1:N} ;\thetavect^{(i-1)}).
\end{align}
It follows that the E-step splits into two steps, an E-W step and an E-Z step. Details of the hybrid-GLLiM algorithm are provided in Appendix~\ref{appendix:em-gllim}. Note that the model described in Section \ref{subsec:invreg_nonhybrid} simply corresponds to $L_w=0$, hence this algorithm can be used to solve both models (fully observed output and partially observed output).  
%A flexible Matlab toolbox implementing these methods is available online, along with head-pose estimation examples.\footnote{https://team.inria.fr/perception/gllim$\_$toolbox/.}

\subsection{Forward Predictive Distribution}
\addnote[label-prob-3]{1}{Once the model parameters $\thetavect$ are estimated, one obtains the low-dimensional to high-dimensional \textit{inverse predictive distribution} as detailed in \cite{DeleforgeForbesHoraud2015}}. 
%\begin{align}
% \label{eq:JGMM_inverse_map}
% p(\yvect|\xvect;\thetavect) &=
%  \sum_{k=1}^K \nu_k \mathcal{N}(\yvect;\Amat_k\xvect+\bvect_k,\Sigmamat_k),\\
%  \textrm{ with } 
%\nu_k &= \frac{\pi_k\mathcal{N}(\xvect ; \cvect_k,\Gammamat_k)}{\sum_{j=1}^K\pi_j\mathcal{N}(\xvect ; \cvect_j,\Gammamat_j)}
%\end{align}
%which is a Gaussian mixture conditioned on the parameters $\thetavect$.
More interesting, it is also possible to obtain the desired high-dimensional to low-dimensional \textit{forward predictive distribution}:
\begin{align}
  \label{eq:JGMM_forward_map}
p(\xvect|\yvect;\thetavect^*)  & = \sum_{k=1}^K \nu_k^*  \mathcal{N}(\xvect;\Amat^*_k\yvect+\bvect^*_k,\Sigmamat_k^*) \\
\label{eq:JGMM_proportions}
\textrm{ with } 
\nu_k^* & = \frac{\pi_k^*\mathcal{N}(\yvect ; \cvect_k^*,\Gammamat_k^*)}{\sum\limits_{j=1}^K\pi_j^*\mathcal{N}(\yvect ; \cvect_j^*,\Gammamat^*_j)} 
\end{align}
which is also a Gaussian mixture conditioned by the parameters $\thetavect^{*}$:
\begin{equation}
\label{eq:thetastar-def}
\thetavect^{*} = \{\cvect_k^{*},\Gammamat_k^{*},\pi_k^{*},\Amat_k^{*}, \bvect_k^{*},\Sigmamat_k^{*}\}_{k=1}^K.
\end{equation}
A prominent feature of this model is that the parameters $\thetavect^{*}$ can be expressed analytically from the parameters $\thetavect$ (please consult the analytical expressions of these parameters in Appendix~\ref{appendix:em-gllim}):
\begin{align}
\label{eq:direct1}
	\cvect_k^* &= \Amat_k\cvect_k + \bvect_k,\\
\label{eq:direct2}
	\Gammamat_k^* &= \Sigmamat_k + \Amat_k\Gammamat_k\Amat_k^\top,\\
	\label{eq:direct3}
	\pi_k^* &= \pi_k,\\
	\label{eq:direct4}
	\Amat_k^* &= \Sigmamat_k^* \Amat_k^\top \Sigmamat_k^{-1},\\
	\label{eq:direct5}
	\bvect_k^* &= \Sigmamat_k^* \left( \Gammamat_k^{-1}\cvect_k - \Amat_k^\top\Sigmamat_k^{-1}\bvect_k\right),\\
	\label{eq:direct6}
	\Sigmamat_k^* &= \left( \Gammamat_k^{-1} + \Amat_k^\top\Sigmamat_k^{-1}\Amat_k \right).^{-1}
\end{align}
Using \eqref{eq:JGMM_forward_map} one can predict an output $\hat{\xvect}$ corresponding to a test input $\hat{\yvect}$
\begin{align}
\label{eq:JGMM_forward_est}
\hat{\xvect} &=	f(\hat{\yvect}) \quad \text{with:}\nonumber \\
f(\yvect) & =	\mathbb{E}\left[\xvect \vert  \yvect; \thetavect^* \right] = \sum_{k=1}^K \nu_k^* \left(\Amat_k^*\yvect + \bvect_k^*\right).
\end{align}

	\section{Implementation Details}
	\label{section:implementation}
	The proposed head-pose estimation method is implemented as follows. We use the Matlab computer vision toolbox implementation of the face detector of \cite{violaJones} as we found that this method yields very good face detections and localizations for a wide range of face orientations, including side views. \addnote[label-imp-1]{1}{The Matlab implementation of \cite{violaJones} offers three different trained classifiers for face detection: two of them for frontal-view detection and one for profile-view detection. These three classifiers yield different results for face detection in terms of bounding-box location and size. The results of face detection using these three classifiers are then  combined together for both training and testing of our method}. For each face detection the associated bounding box is resized to patches of $64\times64$, converted to a grey-level image to which histogram equalization is then applied. A HOG descriptor is extracted from this resized and histogram-equalized patch. To do so, we build a HOG pyramid (p-HOG) by stacking HOG descriptors at multiple resolutions. The following parameters are used to build p-HOG descriptors: 
\begin{itemize}
	\item Block resolution: $2\times2$ cells;
	\item Cell resolutions: $32\times32$, $16\times16$ and $8\times8$, and
	\item Number of orientation bins: $8$
\end{itemize}
Three HOG descriptors are computed, one for each cell resolution, which are then stacked to form a high-dimensional vector $\yvect\in\mathbb{R}^D$, with $D=1888$. 

The dimension of the output variable $\xvect\in\mathbb{R}^L$ depends on the number of pose parameters (up to three angles: yaw, pitch and roll), the bounding-box shift parameters (horizontal and vertical shifts) and the number of latent variables. Hence the output dimension may vary from $L=1$ (one angle, no shift, no latent variable) to $L=10$ (three angles, two shifts, four latent variables). 

We used the algorithm detailed in Appendix~\ref{appendix:em-gllim} to learn the model parameters, either with fully observed output variables (pose angles and bounding-box shifts), or with both observed and latent output variables. From these inverse regression parameters, we derive the forward parameters $\thetavect^{*}$, \ie equations \eqref{eq:direct1}-\eqref{eq:direct6} that allow us to estimate the forward predictive distribution \eqref{eq:JGMM_forward_map}, and hence to predict an output from a test input, \ie \eqref{eq:JGMM_forward_est}.
\addnote[label-impl-1]{2}{
\begin{algorithm}[b!]
\caption{Iterative prediction}
\begin{algorithmic}[1]
\Require Bounding-box location $\uvect$ and forward model parameters $\thetavect^*$
\Procedure{HeadPoseEstimation}{$\uvect$,$\thetavect^*$}
   %\State $i \leftarrow 1$ \Comment{iteration index}
   \Repeat %\Comment{iterative prediction}
      \State Build $\yvect$ from current bounding-box location $\uvect$
      \State Predict $\xvect=[\xvect_{h} ; \xvect_{b}]$ from $\yvect$ using \eqref{eq:JGMM_forward_est}
      \State Update the bounding-box location $\uvect = \uvect + \xvect_{b}$
      %\State $i \leftarrow i+1$
      \Until{$\| \xvect_{b} \| \leq \epsilon$}
   \State \textbf{return} {head-pose $\xvect_{h}$ and bounding-box location $\xvect_{b}$}
\EndProcedure
\end{algorithmic}\label{Alg:iterative}
\end{algorithm}
The joint estimation of the head-pose angles and of the bounding-box shift is achieved iteratively in the following way. The current bounding-box location, $\uvect\in\mathbb{R}^2$, is used to build a feature vector $\yvect$ from which both a head pose $\xvect_{h}$ and a bounding-box shift  $\xvect_{b}$ are predicted. The latter is then used to update the bounding-box location, to build an updated feature vector and to predict an updated head pose and a new bounding-box shift. This iterative prediction is described in detail in Algorithm~\ref{Alg:iterative}}.

%helps to compensate the misalignment of the face detection, running an iterative scheme that will refine the face position at each iteration using the offset prediction and build a new descriptor $\yvect$ will make the face position converge to an optimal localization that will reduce the misalignment. Algorithm~\ref{Alg:iterative} describe the iterative formulation, we define the face position as a vector $\uvect$ and the bounding-box offset prediction from $\xvect$ as $\xvect_{bb}$.

%\textbf{CHANGE THIS PARAGRAPH WITH MORE DETAILED REFERENCES TO EQUATIONS AND ALGORITHMS.} The estimation of the output vector $\xvect$ is done using Equation~\ref{eq:JGMM_forward_est} where $\yvect$ is the associated face feature information. In the case where $\xvect$ contains information about the bounding box offset the initial face position is refined using the offset estimation, and new face feature is computed using this update face position, another iteration of the estimation of $\xvect$ is done. We have an iterative estimation of the head pose and face bounding box offset.

	\section{Experiments}
	\label{section:experiments}
	In this section we present an experimental evaluation of the proposed head-pose estimation methodology. The experiments are carried out with three publicly available datasets:  the Prima dataset \cite{gourier2004estimating}, the BIWI dataset \cite{fanelli2013random}, and the McGill real-world face video dataset \cite{demirkus2013robust,demirkus2015hierarchical}: %Sample images from the different datasets are displayed in in Fig.~\ref{fig:database}.
\begin{itemize}
\item The \textbf{Prima head pose dataset} consists of 2790 images of 15 persons recorded twice. Pitch values lie in the interval $[-60^{\circ},60^{\circ}]$, and yaw values lie in the interval $[-90^{\circ},90^{\circ}]$  with a $15^{\circ}$ step. Thus, there are 93 poses available for each person. All the recordings were achieved with the same background. One interesting feature of this dataset is the the pose space is uniformly sampled. The dataset is annotated such that a face bounding box (manually annotated) and the corresponding yaw and pitch angle values are provided for each sample. 

\item The \textbf{Biwi Kinect head pose dataset} consists of video recordings of 20 people (16 men, 4 women, some of them recorded twice) using a Kinect camera. During the recordings, the participants freely move their head and the corresponding head angles lie in the intervals $[-60^{\circ},60^{\circ}]$ (pitch), $[-75^{\circ},75^{\circ}]$ (yaw), and $[-20^{\circ},20^{\circ}]$ (roll). Unlike the Prima dataset, the parameter space is not evenly sampled. The face centers (nose tips) were detected on each frame in the dataset, which allowed us to automatically extract a bounding box for each sample.

\item The \textbf{McGill real-world face video dataset} consists of 60 videos (a single participant per video, 31 women and 29 men) recorded with the objective of studying unconstrained face classification. The videos were recorded in different environments (both indoor or outdoor) thus resulting in arbitrary illumination conditions and background clutter. Furthermore, the participants were completely free in their behaviors and movements. 
Yaw angles range in the interval $[-90^{\circ},90^{\circ}]$. Yaw values corresponding to each video frame are estimated using a two-step labeling procedure that provides the most likely angle as well as a degree of confidence. The labeling consists of showing images and possible angle values to human experts, \ie \cite{demirkus2013robust}.
\end{itemize}
%\begin{figure*}[htb]
%	\centering
%	\includegraphics[width=\textwidth]{figures/database.png}
%	\caption{Examples of face images from the 3 used datasets, Left Prima, Middle Biwi-Kinect and Right McGill}
%	\label{fig:database}
%\end{figure*}
We adopted the leave-one-out evaluation protocol at the individual person level. More precisely, all the images/frames associated with one participant are left aside and used for testing, while the remaining data were used for training. As a measure of performance, we evaluated the absolute error between an estimated angle and the ground-truth value, then we computed the \textit{mean absolute error} (MAE) and standard deviation (STD) over several tests.

We experimented with the following variants of the proposed algorithm (notice that the number of pose angles depends on the dataset, \ie yaw, pitch and roll (BIWI), yaw and pitch (Prima) or yaw (McGill)):
\begin{itemize}
\item {\it GLLiM\_pose} learns and predicts with one, two, or three pose angles;
\item {\it hGLLiM\_pose-d} learns with the pose parameters as well as with partially latent output, where the number $d$ of latent variables varies between 1 and 4;
\item {\it GLLiM\_pose\&bb} learns and predicts with both pose angles and bounding-box shifts, and
\item  {\it hGLLiM\_pose\&bb-d} learns and predicts with pose angles, bounding-box shifts and partially latent output.
\end{itemize}

\addnote[face-alignment]{1}{An important aspect of any head-pose method is the way faces are detected in images. We used manually annotated bounding boxes, whenever they are available with the datasets. Otherwise, we used bounding boxes provided with a face detector, \eg \cite{violaJones}. To evaluate the robustness to inaccurate face localization, we introduced random shifts, drawn from a Gaussian distribution, and we used these shifts in conjunction with {\it GLLiM\_pose\&bb} and with {\it hGLLiM\_pose\&bb-d} to learn the regression parameters and to predict the correct bounding-box location. Notice that in the case of the latter algorithms, the prediction is run iteratively, i.e. Algorithm~\ref{Alg:iterative}: Extract a HOG vector, predict the pose and the shift, extract a HOG vector from the shifted bounding box, predict the pose and the shift, etc. This is stopped when the shift becomes very small and, as it can be seen below, it considerably improves the quality of the head-pose method.}

\begin{figure}[htb]
	\includegraphics[width=\columnwidth]{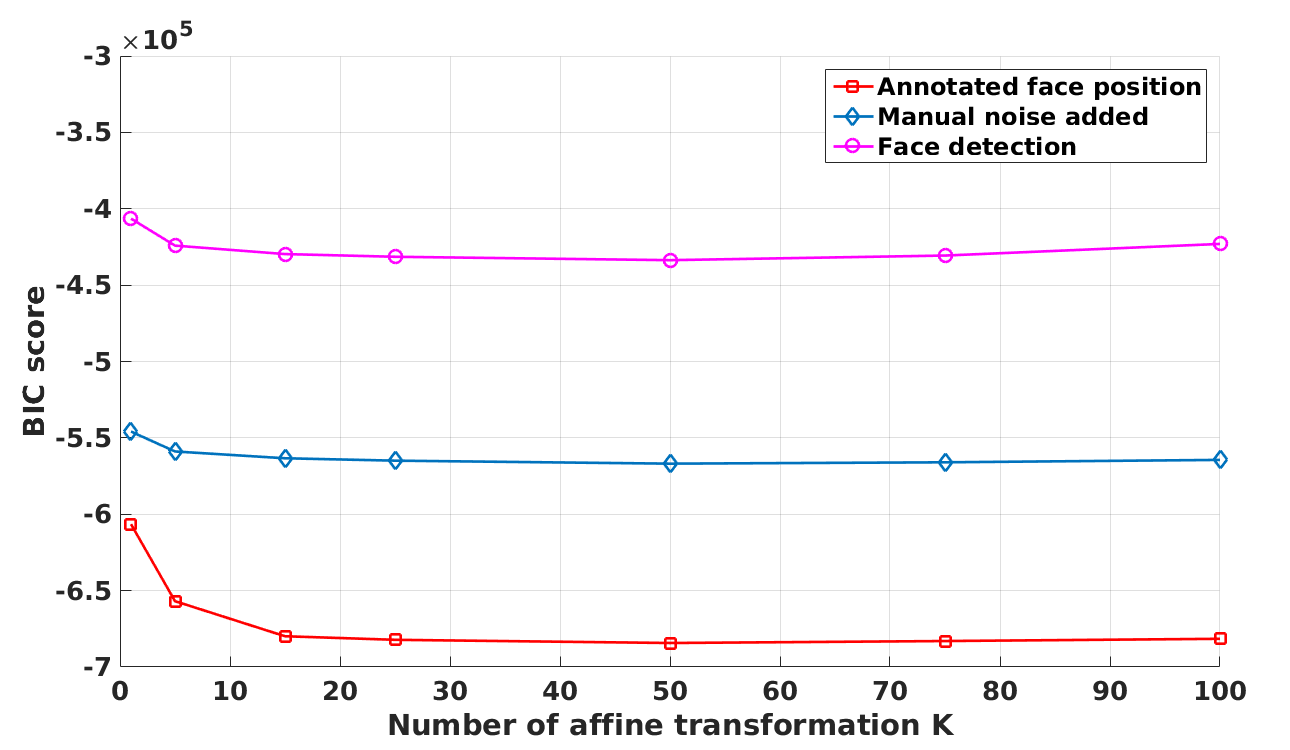}
	\caption{The Bayesian information criterion (BIC) as a function of the number of affine transformations in GLLiM. These experiments use the Prima dataset with the leave-one-out protocol.}
	\label{fig:bic}
\end{figure}

%\def\arraystretch{2}
%\begin{table}[!h]
%	\centering
%	\caption{\label{table:BIC} Bayesian Information Criterion (BIC) score several models learned with different values of $K$ on the Prima dataset. We use GLLiM\_pose to learn each model with different input data, using Annotated Face position (AFP), Adding manual noise to the face position (MNA) and using a Face Detector (FD), The best results are in bold.}
%	\begin{tabular}{l|| c c c c c c c}
%		& \multicolumn{7}{c}{Bayseian Information Criterion Score}\\
%		K & $1$  & $5$ & $15$ & $25$ & $50$ & $75$ & $100$ \\
%	\hline
%		AFP & $-6.0608$ & $-6.5845$ & $-6.7986$ & $-6.822$  & $\mathbf{-6.8429}$ & $-6.8228$ & $-6.8173$\\
%		MNA & $-5.4554$ & $-5.6018$ & $-5.6335$ & $-5.6491$ & $\mathbf{-5.6688}$ & $-5.655$  & $-5.6455$\\
%		FD  & $-4.0596$ & $-4.2602$ & $-4.2967$ & $-4.3144$ & $\mathbf{-4.3366}$ & $-4.3049$ & $-4.2307$
%	\end{tabular}
%	%\vspace*{0.3cm}
%\end{table}
\def\arraystretch{2}
\begin{table}[!h]
	\centering
	\caption{\label{table:BIC} The BIC score for several models learned with different values of $K$ using the Prima dataset. We use GLLiM\_pose to learn each model with different input data (Fig.~\ref{fig:bic}):  annotated face position (AFP), adding manual noise to the face position (MNA) and using a face detector (FD), The optimal BIC scores are in bold.}
	\begin{tabular}{l|| c c c c c}
		%& \multicolumn{5}{c}{Bayseian Information Criterion Score}\\
		Data & $K=1$  & $K=5$ & $K=25$ & $K=50$ & $K=100$ \\
	\hline
		AFP & $-6.0608$ & $-6.5845$ & $-6.822$  & $\mathbf{-6.8429}$ & $-6.8173$\\
		MNA & $-5.4554$ & $-5.6018$ & $-5.6491$ & $\mathbf{-5.6688}$ & $-5.6455$\\
		FD  & $-4.0596$ & $-4.2602$ & $-4.3144$ & $\mathbf{-4.3366}$ & $-4.2307$
	\end{tabular}
	%\vspace*{0.3cm}
\end{table}
		 
%\begin{figure}[htb]
%	\includegraphics[width=\columnwidth]{figures/bic.png}
%	\caption{Bayesian Information Criterion value as a function of the number of affine transformations in the mixture of linear regression model. These experiments use the Prima dataset with the leave-one-out protocol.}
%	\label{fig:bic}
%\end{figure}

\begin{figure}[htb]
	\includegraphics[width=\columnwidth]{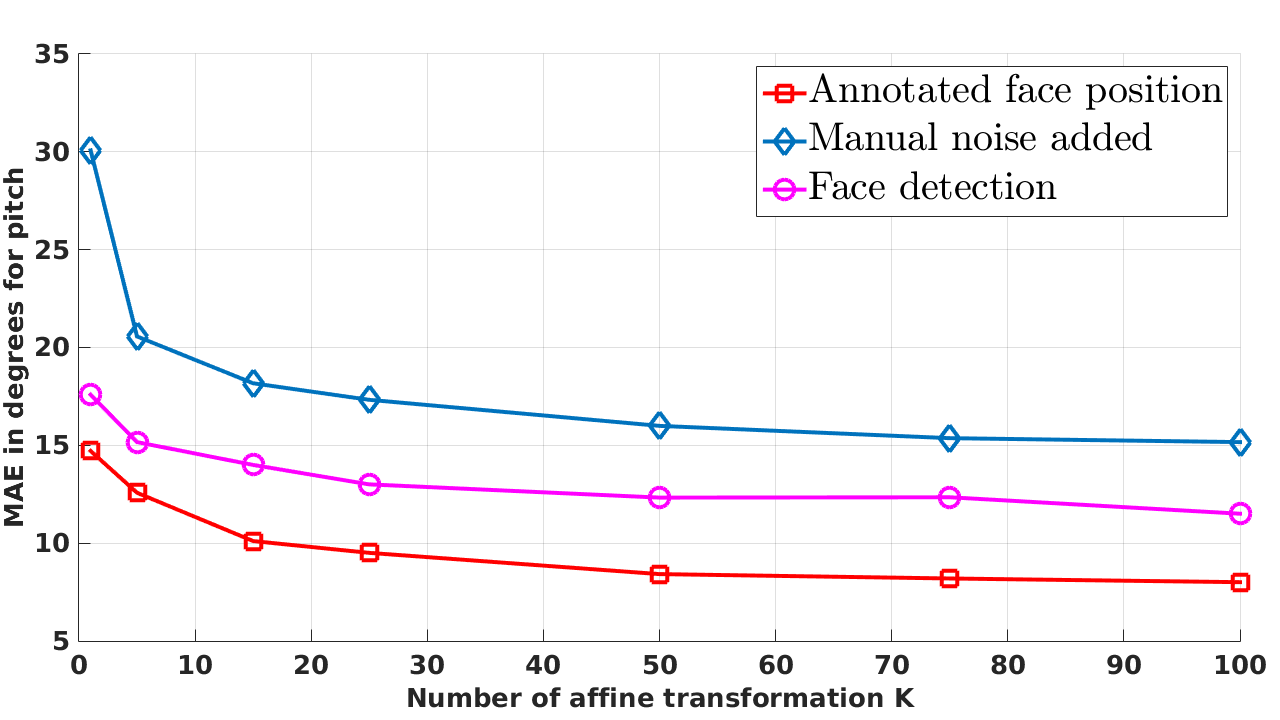}
	\includegraphics[width=\columnwidth]{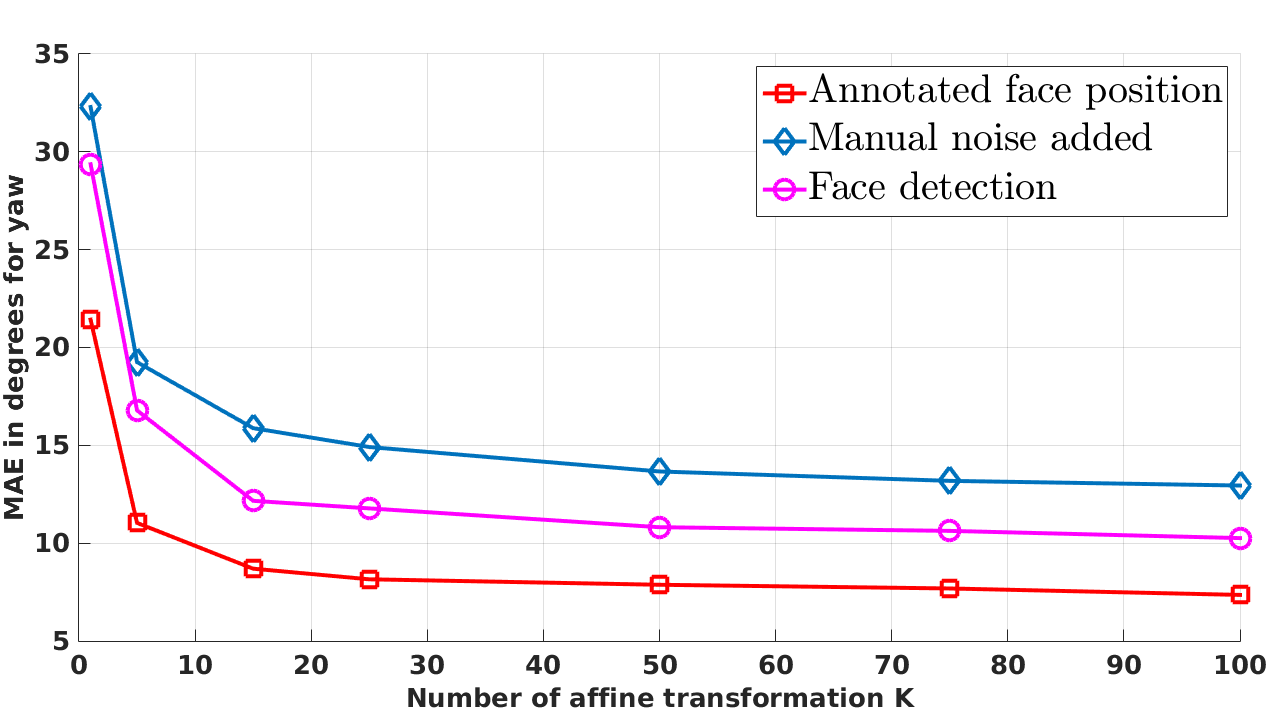}	
	\caption{Mean absolute error (MAE) in degrees, for pitch (top) and yaw (bottom), as a function of the number of affine transformations in the mixture of linear regression model. We used GLLiM\_pose to learn the model parameters independently for pitch and yaw. The three curves correspond to the following face detection cases: manual annotation (red curve), manual annotations with additive noise (blue), and automatic face detection (magenta). These experiments use the Prima dataset with the leave-one-out protocol.}
	\label{fig:noiseDetection}
\end{figure}
\def\arraystretch{2}
\begin{table*}[p]
	\centering
	\caption{\label{table:errorPrima} Mean absolute error (MAE) and standard deviation (STD) (in degrees) obtained with various head-pose methods, regression methods, and our method using the Prima dataset. This dataset contains manually annotated bounding boxes of faces and the corresponding pitch and yaw angles. In order to test the robustness we simulated shifted bounding boxes. The best results are in bold.}
	\begin{tabular}{l|| c c | c c || c c | c c }
		& \multicolumn{4}{c||}{Manually annotated bounding boxes} & \multicolumn{4}{c}{Manual annotation + simulated shifts} \\
		& \multicolumn{2}{c|}{Pitch} & \multicolumn{2}{c||}{Yaw} & \multicolumn{2}{c|}{Pitch} & \multicolumn{2}{c}{Yaw} \\
		Method  & MAE & STD & MAE & STD & MAE & STD & MAE & STD \\
	\hline
		Stiefelhagen \cite{stiefelhagen2004estimating}$^{\ddag}$& $9.7$ & - & $9.5$ & - & - & - & - & - \\	
		Gourier et al. \cite{gourier2007head}$^{\ddag}$& $12.1$ & - & $7.3$ & - & - & - & - & - \\	
		GPR \cite{Rasmussen06gaussianprocesses}$^{\dag}$ & $11.94$ & $10.19$ & $15.04$ & $12.24$ & $19.96$ & $16.58$ & $23.69$ & $18.16$ \\
		PLS \cite{abdi2003partial}$^{\dag}$ & $12.25$ & $9.73$ & $13.38$ & $10.8$ & $17.77$ & $14.47$ & $17.34$ & $13.94$ \\
		SVR \cite{smola2004tutorial}$^{\dag}$ & $11.25$ & $9.42$ & $12.82$ & $10.99$ & $17.09$ & $14.81$ & $17.27$ & $14.09$ \\
		GLLiM\_pose    & $8.41$  & $10.65$ & $7.87$  & $8.08$  & $15.99$ & $16.69$ & $13.66$ & $14.78$ \\
		hGLLiM\_pose-2 & $\mathbf{8.47}$ & $\mathbf{10.35}$ & $\mathbf{7.93}$ & $\mathbf{7.9}$ &  $12.64$ & $14.49$ & $11.51$ & $11.37$ \\
		hGLLiM\_pose-4 & $8.5$  & $10.8$ & $7.85$ & $7.98$ & $\mathbf{12.03}$ & $13.92$ & $\mathbf{10.78}$ & $9.77$ \\
		GLLiM\_pose\&bb    & - & - & - & - & $13.13$ & $13.65$ & $11.3$  & $10.55$ \\
		hGLLiM\_pose\&bb-2 & - & - & - & - & $12.52$ & $\mathbf{12.44}$  & $11.04$ & $9.7$  \\
		hGLLiM\_pose\&bb-4 & - & - & - & - & $12.12$ & $12.85$ & $11.27$ & $\mathbf{9.53}$ \\	
	\end{tabular}
	%\vspace*{0.3cm}
\end{table*}

\def\arraystretch{2}
\begin{table*}[p]
	\centering
	{
	\caption{\label{table:errorBiwi} The mean absolute error (MAE) and standard deviation (STD), expressed in degrees, obtained with various head-pose methods, regression methods, and our method using the BIWI dataset. This dataset contains annotated bounding boxes of faces and the corresponding pitch, yaw, and roll angle values. In order to test the robustness we simulated shifted bounding boxes both for training and for testing. The best results are in bold. Note that \cite{fanelli2013random} uses depth data only and \cite{wang2013head} and \cite{ghiass2015highly} use both color and depth information. Papers using depth data are marked with a $^*$.
	}
	\begin{tabular}{l||c c|c c|c c||c c|c c|c c}
	& \multicolumn{6}{c||}{Manually annotated bounding boxes} & \multicolumn{6}{c}{Manual annotation + simulated shifts} \\
	& \multicolumn{2}{c|}{Pitch} & \multicolumn{2}{c|}{Yaw} & \multicolumn{2}{c||}{Roll} & \multicolumn{2}{c|}{Pitch} & \multicolumn{2}{c}{Yaw} & \multicolumn{2}{c}{Roll} \\
	Method  & MAE & STD & MAE & STD & MAE & STD & MAE & STD & MAE & STD & MAE & STD \\
	\hline
		Ghiass et al. \cite{ghiass2015highly}$^{\ddag *}$ &$\mathbf{0.1}$ & $6.7$ & $\mathbf{0.2}$ & $8.7$ & $\mathbf{0.3}$ & $9.3$ & - & - & - & - & - & - \\
		Fanelli et al. \cite{fanelli2013random}$^{\ddag *}$& $3.8$ & $6.5$ & $3.5$ & $6.8$ & $5.4$ & $6.0$ & - & - & - & - & - & - \\
		Wang et al. \cite{wang2013head}$^{\ddag *}$& $8.5$ & $11.1$ & $8.8$ & $14.3$ & $7.4$ & $10.8$ & - & - & - & - & - & - \\
		GPR \cite{Rasmussen06gaussianprocesses}$^{\dag}$ & $9.64$ & $8.85$ & $7.72$ & $7.17$ & $6.01$ & $6.29$ & $10.77$ & $9.45$ & $9.06$ & $8.33$ & $6.54$ & $6.72$ \\
		PLS \cite{abdi2003partial}$^{\dag}$ & $7.87$ & $6.73$ & $7.35$ & $6.06$ & $6.11$ & $5.9$ & $10.32$ & $8.64$ & $8.67$ & $7.7$ & $6.69$ & $6.74$ \\
		SVR \cite{smola2004tutorial}$^{\dag}$ & $7.77$ & $6.85$ & $6.98$ & $6.26$ & $5.14$ & $5.96$ & $10.82$ & $9.22$ & $9.14$ & $8.32$ & $6.26$ & $7.16$\\
		GLLiM\_pose & $5.77$ & $5.77$ & $4.48$ & $\mathbf{4.33}$ & $4.71$ & $5.31$ & $11.33$ & $11.58$ & $10.2$ & $11.34$ & $7.76$ & $8.02$ \\
		hGLLiM\_pose-2 & $5.57$ & $5.48$ & $4.33$ & $4.68$ & $4.37$ & $5.09$ & $9.04$ & $9.13$ & $7.65$ & $8.3$ & $6.3$ & $6.75$ \\
		hGLLiM\_pose-4 & $5.43$  & $\mathbf{5.44}$  & $4.24$ & $5.37$  & $4.13$  & $\mathbf{4.86}$ & $8.45$  & $8.41$  & $6.93$ & $7.72$  & $6.12$  & $6.8$\\
%	\hline
		GLLiM\_pose\&bb    & - & - & - & - & - & - & $8.49$ & $8.79$ & $6.86$  & $7.3$ & $6.57$ & $6.95$ \\
		hGLLiM\_pose\&bb-2 & - & - & - & - & - & - & $7.81$ & $7.68$ & $6.41$ & $7.19$ & $5.75$ & $6.68$ \\
		hGLLiM\_pose\&bb-4 & - & - & - & - & - & - & $\mathbf{7.65}$ & $\mathbf{8.0}$ & $\mathbf{6.06}$ & $\mathbf{6.91}$ & $\mathbf{5.62}$ & $\mathbf{6.35}$ \\
	\end{tabular}
	}
%	\vspace*{0.3cm}
\end{table*}

The number $K$ of Gaussian components is an important parameter, which in our model corresponds to the number of affine mappings. We carried out several experiments in order to evaluate the quality of the results obtained by our method as a function of the number of affine transformations in the mixture. We use the GLLiM\_pose variant of our algorithm with three different face detection options: manual annotation, manual annotation perturbed with additive Gaussian noise, and automatic face detection. We trained these three versions of GLLiM\_pose with $K$ varying from 1 to 100. 

\addnote[label-exp-1]{2}{In order to determine the optimal number of affine mappings, $K$, associated with GLLiM, we use two criteria, the Bayesian information criterion (BIC) which is an information-theoretic criterion generally used for model selection, and an experimental figure of merit based on the mean absolute error (MAE). We learned several models for different values of $K$ using the Prima dataset. We seek the model that yields low BIC and MAE scores. The BIC and MAE values are plotted as a function of $K$ in Table~\ref{table:BIC} and in Fig.~\ref{fig:bic} and Fig.~\ref{fig:noiseDetection}. These curves show the same behavior: as the number of affine mappings increases from $K=1$ to approximatively $K=30$, both the BIC and MAE scores decrease, then the curve slopes become almost horizontal. Both BIC and MAE reach the lowest score for $K=50$. This behavior can be explained as follows. When $K<5$ the model is not flexible enough to take into account the apparently non-linear mapping between HOG features and head-pose parameters. It can be observed from  Fig.~\ref{fig:noiseDetection} that a large value for $K$ increases the model accuracy. As expected, the computational complexity increases with $K$ as well. Indeed, the number of model parameters is linear in the number of mixture components and hence the size of the training dataset must be increased as well. It is well known that a large number of components in a mixture model presents the risk of overfitting. It is interesting to notice that BIC (derived from information theory) and MAE (based on experiments with the data) yield the same optimal value, namely $K\approx 50$.}

% Fig.~\ref{fig:noiseDetection} shows the performance of head-pose estimation as a function of $K$ for the Prima Dataset. As it can be seen on this figure, the estimation error decreases as the number of affine mappings increases from  K=1 to K=30. Then, from $K=30$ the slope of the plot becomes almost horizontal, with a plateau beyond $K=50$. This behavior can be explained as follows. For $K<5$, the model is not flexible enough to take into account the apparently non-linear mapping between HOG features and head-pose parameters. It can be observed from  Fig.~\ref{fig:noiseDetection} that a large value for $K$ increases the model accuracy. As expected, the computational complexity increases with $K$ as well. Indeed, the number of model parameters is linear in the number of mixture components and hence the size of the training dataset must be increased as well. For all these reasons, in all the reported experiments, we fixed the number of affine transformation to  $K=50$.

The proposed algorithms were compared with the following state-of-the-art head-pose estimation methods: the neural-network based methods of \cite{stiefelhagen2004estimating}, \cite{gourier2007head} and of \cite{ahn2014real} , the method of \cite{demirkus2012soft} based on dictionary learning, the graphical-model method of \cite{demirkus2014probabilistic}, the template based method of \cite{zhu2012face}, the supervised non-linear optimization method of \cite{xiong2013supervised}, the optimization method of \cite{ghiass2015highly}, and the random-forest methods of \cite{fanelli2013random} and of \cite{wang2013head}. Additionally, we benchmarked the following regression methods: support vector regression (SVR) \cite{smola2004tutorial}, Gaussian process regression (GPR) \cite{Rasmussen06gaussianprocesses}, and partial least squares (PLS) \cite{abdi2003partial}, as they are widely known and commonly used regression methods for which software packages are publicly available. Notice that some of these methods estimate only one parameter, \ie the yaw angle \cite{demirkus2012soft,demirkus2014probabilistic,zhu2012face,xiong2013supervised}, while the random-forest methods of \cite{fanelli2013random}, \cite{ghiass2015highly} and \cite{wang2013head} use depth information available with the BIWI (Kinect) dataset. 
%%%%%%%%%%%%%%%%%% TABLES FOR RESULTS

Table \ref{table:errorPrima}, Table \ref{table:errorBiwi}, and Table \ref{table:errorMcgill} show the results of head-pose estimation obtained with the Prima, BIWI, and McGill datasets, respectively. \addnote[label-exp-3]{1}{The $^{\dag}$ symbol indicates that the results are those reported by the authors while the $^{\ddag}$ symbol indicates that the results are obtained using either publicly available software packages or our own implementations}. 
In the case of the Prima dataset, GLLiM\_pose and hGLLiM\_pose yield the best results. We note that hGLLiM\_pose\&bb variants of the algorithm (simultaneous prediction of pose, bounding-box shift and partially-latent output) increase the confidence (low STD).
\cref{table:errorBiwi} shows the results obtained with the BIWI datasets. As already mentioned,\addnote[label-exp-1]{1}{\cite{fanelli2013random} uses depth information and \cite{wang2013head}, \cite{ghiass2015highly} use of depth and color information}. Overall, the proposed algorithms compare favorably with \cite{fanelli2013random}. hGLLiM\_pose-4 yields the best MAE for the roll angle, while \addnote[label-exp-2]{1}{\cite{ghiass2015highly} yields the best MAE for pitch and yaw, but with a high standard deviation}. Our algorithm estimates the parameters with the highest confidence (lowest standard deviation). \cref{table:errorMcgill} shows the results obtained with the McGill dataset. The ground-truth yaw values in this dataset are obtained by human experts that must choose among a discrete set of 7 values. Clearly, this is not enough to properly train our algorithms. The method of \cite{demirkus2014probabilistic} yields the best results in terms of RMSE while hGLLiM\_pose-2 yields the best results in terms of MAE. Notice that PLS yield the highest confidence in this case.

\def\arraystretch{2}
\begin{table}[!h]
%	\centering{
		\caption{\label{table:errorMcgill} Root mean square error (RMSE), mean absolute error (MAE) and standard deviation (STD) (in degrees) obtained with various head-pose methods, regression methods, and our method using the McGill Real-World dataset. This dataset contains annotated yaw angles. Bounding boxes are located with a face detector. The best results are in bold.}
	\begin{tabular}{l||c c c}
		& \multicolumn{3}{c}{Bounding boxes based on face detection} \\
		& \multicolumn{3}{c}{Yaw} \\
		Method  & RMSE & \hspace*{0.7cm}MAE & STD\\
	\hline
		Demirkus et al. \cite{demirkus2012soft}$^{\ddag}$ & $> 40$ & \hspace*{0.7cm}- & - \\
		Xiong and De la Torre \cite{xiong2013supervised}$^{\ddag}$ & $29.81$ & \hspace*{0.7cm}- & -\\ 
		Zhu and Ramanan \cite{zhu2012face}$^{\ddag}$& $35.70$ & \hspace*{0.7cm}- & - \\
		Demirkus et al. \cite{demirkus2014probabilistic}$^{\ddag}$ & $\mathbf{12.41}$ & \hspace*{0.7cm}- & - \\
		GPR \cite{Rasmussen06gaussianprocesses}$^{\dag}$ & $23.18$ & \hspace*{0.7cm}$16.22$ & $16.71$ \\
		PLS \cite{abdi2003partial}$^{\dag}$ & $22.46$ & \hspace*{0.7cm}$15.56$ & $\mathbf{16.2}$ \\
		SVR \cite{smola2004tutorial}$^{\dag}$ & $21.13$ & \hspace*{0.7cm}$15.25$ & $18.43$ \\
		GLLiM\_pose & $26.62$ & \hspace*{0.7cm}$13.1$ & $23.17$ \\
		hGLLiM\_pose-2 & $24.0$ & \hspace*{0.7cm}$\mathbf{11.99}$ & $20.79$ \\ 
		hGLLiM\_pose-4 & $24.25$ & \hspace*{0.7cm}$12.01$ & $21.06$ \\
	\end{tabular}
%	}
	\vspace*{0.3cm}
\end{table} 

\begin{figure}[htb]
	\includegraphics[width=\columnwidth]{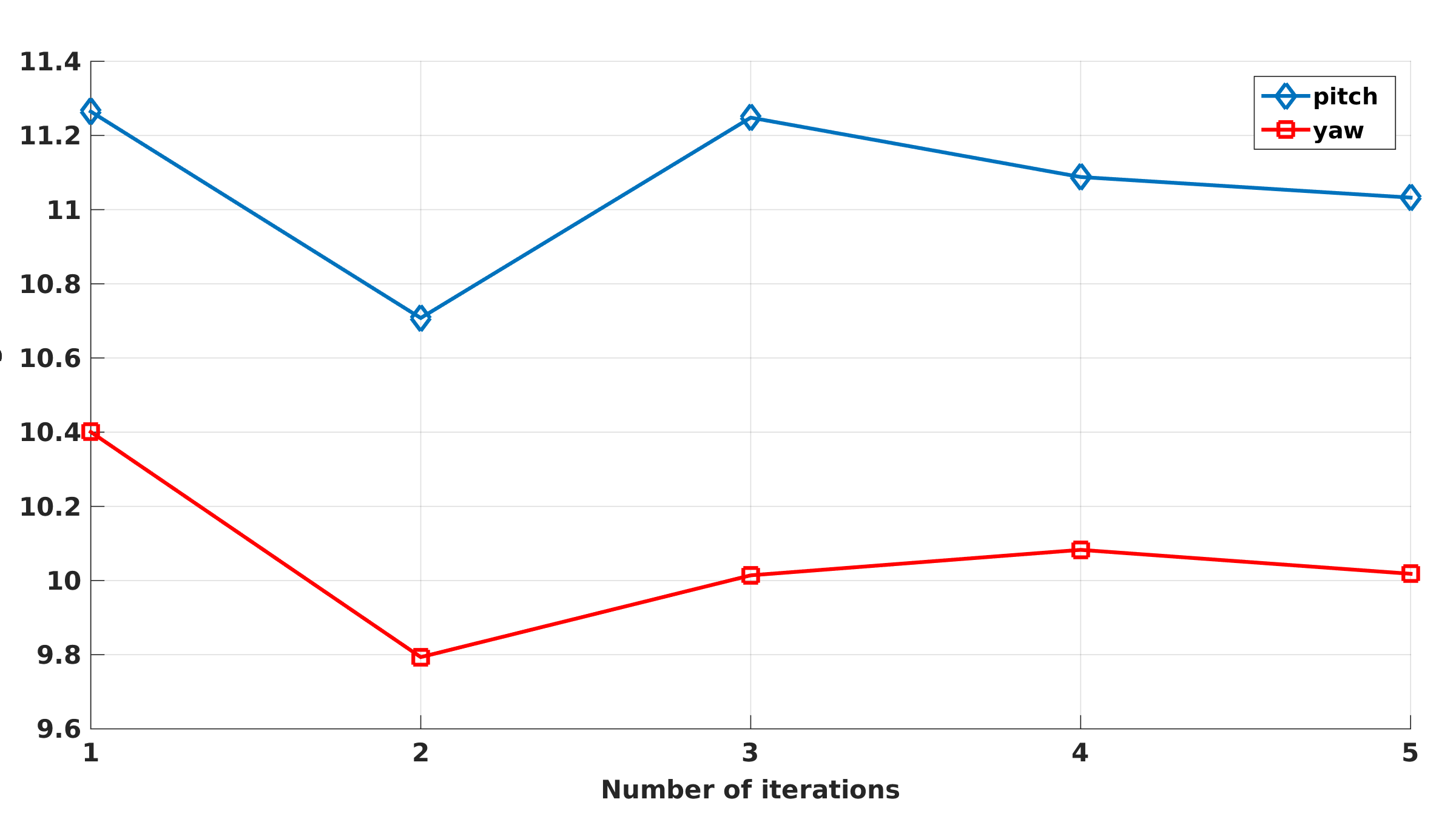}
	\caption{Mean absolute error (MAE) for pitch and yaw as a function of the number of iterations of GLLiM\_pose\&bb.}
	\label{fig:ite}
\end{figure}

%\begin{figure*}[htb]
%	\includegraphics[width=\textwidth]{figures/examples.png}
%	\caption{Some examples of head pose estimation results, top line comparing gllim\_pose (red arrow) and hGLLiM\_pose (green arrow), bottom the face bounding box refinement, original detected face bounding box in red and refined face bounding box in green}
%	\label{figure:examples}
%\end{figure*}

The results summarized in \cref{table:errorPrima}, \cref{table:errorBiwi} and \cref{table:errorMcgill} allow to quantify the variants of the proposed algorithm. With manually annotated bounding boxes, \eg \cref{table:errorPrima,table:errorBiwi}, there is no notable difference between these variants. Whenever the regression is trained with simulated bounding-box shifts, both GLLiM\_pose\&bb and hGLLiM\_pose perform better than GLLiM\_pose. It is interesting to note that in the case of the Prima dataset (\cref{table:errorPrima}), hGLLiM\_pose-4 (the dimension of the latent part of the output is 4) is the best-performing variant of GLLiM, while in the case of the BIWI dataset (\cref{table:errorBiwi}), hGLLiM\_Pose\&bb-4 is the best performing one. We also experimented with a larger latent dimension without improving the accuracy. Concerning the McGill dataset, neither GLLiM\_pose\&bb nor hGLLiM\_pose\&bb could be trained because the ground-truth face bounding boxes are not available. In the case of this dataset, we used a face detector both for training and for testing. The fact that, overall, hGLLiM\_pose performs better than GLLiM\_pose validates the advantage of adding a latent component to the output variable. The latter ``absorbs" various phenomena that would otherwise affect the accuracy of the pose parameters.

As already mentioned, GLLiM\_pose\&bb and hGLLiM\_pose\&bb are applied iteratively, until there is no improvement in the predicted output. Fig.~\ref{fig:ite} plots the MAE of yaw and pitch as a function of the number of iterations of GLLiM\_pose\&bb. As it can be observed from these curves, the MAE decreases after two iterations and then it slightly increases and again it decreases. Therefore, in practice we run two iterations of these algorithms.

\begin{figure*}[htb]
\centering
\begin{tabular}{c}
	\includegraphics[height=0.35\textwidth]{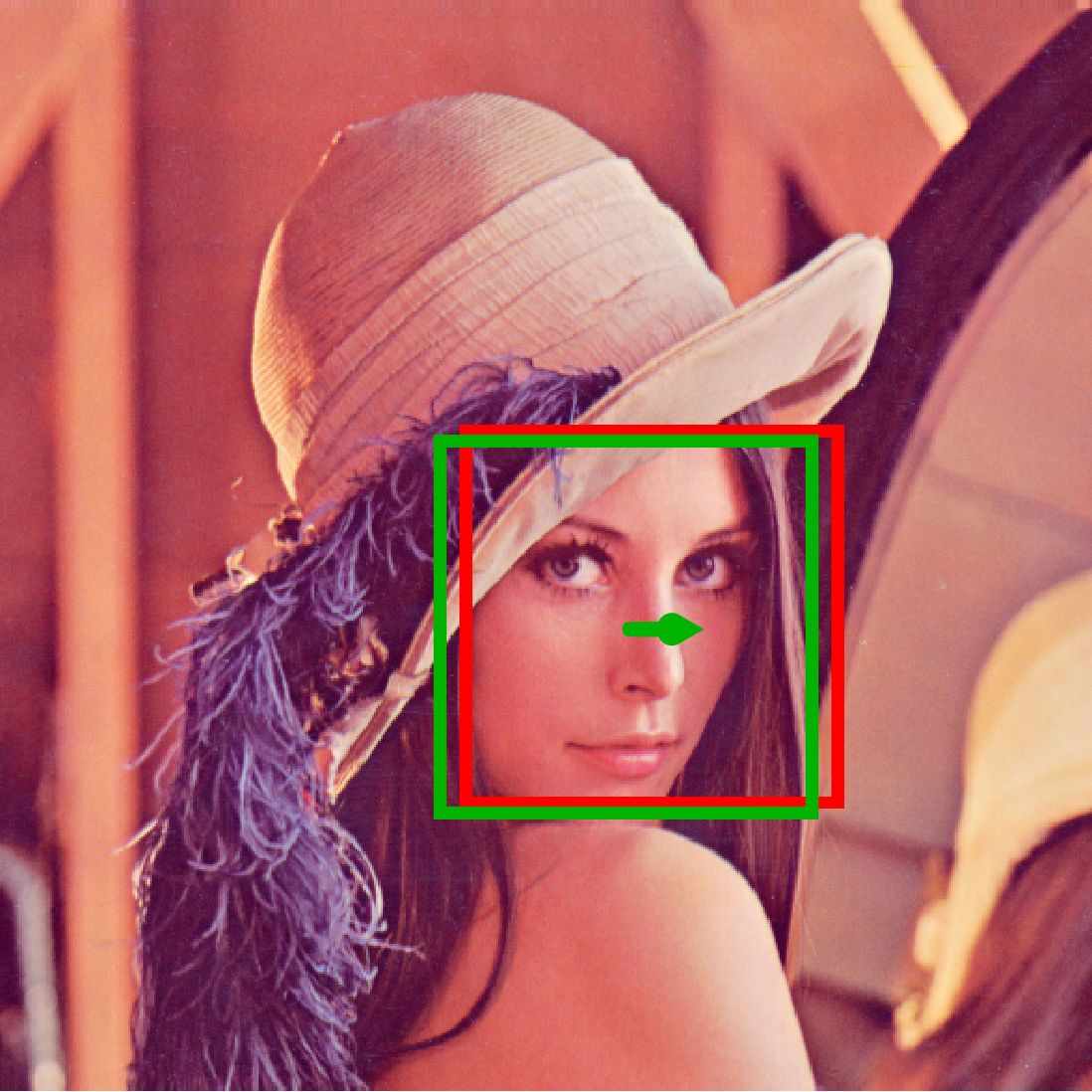}
	\includegraphics[height=0.35\textwidth]{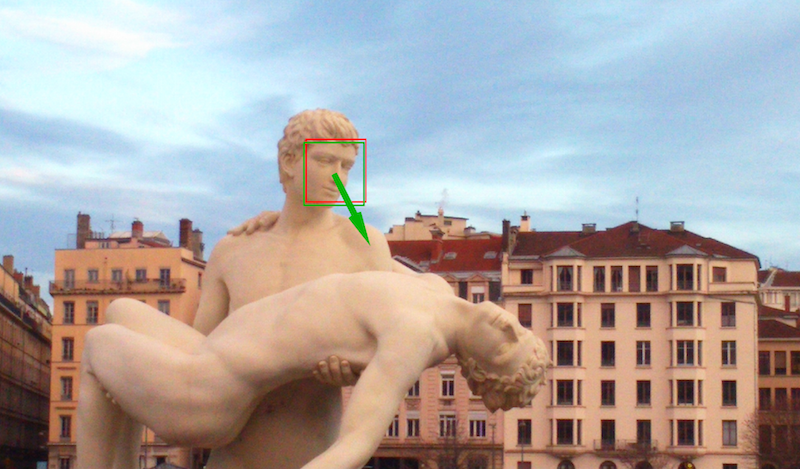}\\
	\includegraphics[height=0.295\textwidth]{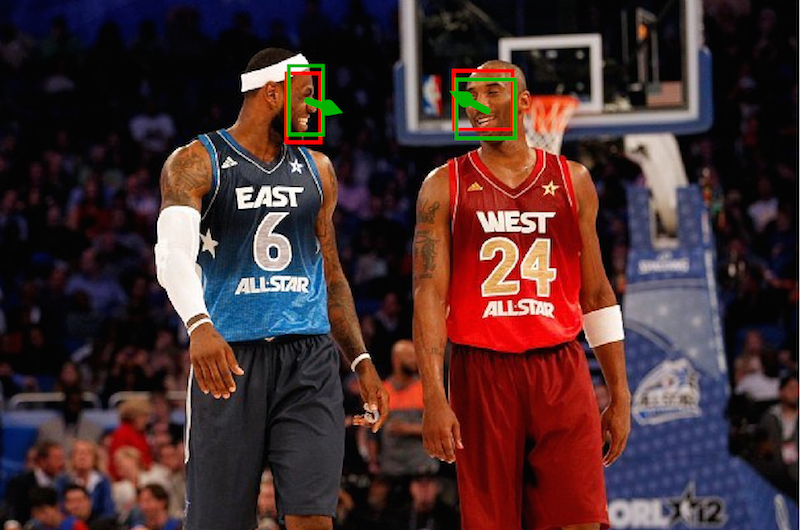}
	\includegraphics[height=0.295\textwidth]{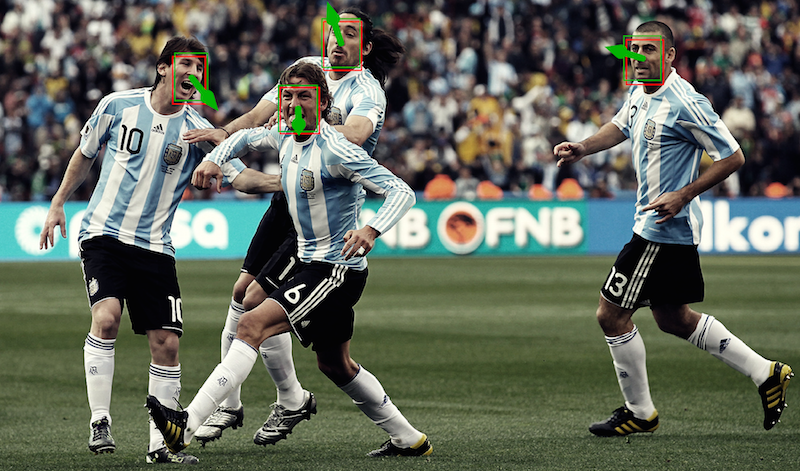}
	\end{tabular}
	\caption{Examples of simultaneous estimation of head-pose angles and of bounding-box shifts. The initial bounding box (found with an automatic face detector) is shown in red. The estimated bounding box is shown in green.}
	\label{figure:examples}
\end{figure*}

Finally, we applied hGLLiM\_Pose\&bb-4 to a set of images that are not contained in any of the three datasets. Fig.~\ref{figure:examples} shows some of these results. The face detector of \cite{violaJones} was used to detect faces (red bounding boxes). The output of hGLLiM\_Pose\&bb-4: the shifted bounding boxes (shown in green) and the estimated pitch, roll, and yaw angles (shown with a green arrow).

%Some examples of the iterative face bounding box refinement and head pose estimation, and also head pose estimation using latent and no latent variables are shown in Fig.~\ref{figure:examples}.

	\section{Conclusions}
	\label{section:conclusion}
	In this paper we proposed a solution to the problem of estimating head pose from the bounding box aligned with a face. Instead of extracting facial landmarks, the method directly maps high-dimensional feature vectors (extracted from faces) onto a low-dimensional manifold. The method relies on learning a mixture of linear regressions. The latter is modeled within the framework of generative methods. More precisely, it is assumed that the high-dimensional feature space is generated from a low-dimensional parameter space. Consequently, an inverse regression strategy is adopted: a low-dimensional to high-dimensional regression is learnt, followed by Bayes inversion. 

We experimented with four variants of the proposed algorithm: (i)~GLLiM\_pose, which learns and predicts the pose parameters, (ii)~hGLLiM\_pose, which learns and predicts the pose parameters in the presence of latent variable augmentation of the output, (iii)~GLLiM\_pose\&bb, which learns and predicts both the pose parameters and bounding-box shifts, and (iv)~hGLLiM\_pose\&bb which combines hGLLiM with GLLiM\_pose\&bb. The experiments and benchmarks, carried out with three publicly available datasets, show
that the latent-augmentation variants of the algorithm improve the accuracy of the estimation and perform better than several state-of-the-art algorithms. 

The methodology presented in this paper has not been tuned for the particular application of head-pose estimation. The algorithms may be applied, with minor modifications, to other high-dimensional to low-dimensional mapping problems, \eg estimation of human pose from color and depth images. It may also be used as input for gaze estimation or for determining the visual focus of attention, e.g. \cite{masse:hal-01301766}.

In the future we plan to extend the method to the problem of tracking pose parameters. Indeed, a natural extension of the proposed method is to incorporate a dynamic model to better predict the output variables over time. Such a model could be simultaneously applied to the pose parameters and to the bounding-box shifts. Hence, one can track the image region of interest and the pose parameters in a unified framework.
	\appendices
	\section{EM for GLLiM and Hybrid-GLLiM}
	\label{appendix:em-gllim}
	This appendix details the EM algorithm that estimates the parameters of the regression method described in Section~\ref{section:probabilistic}. The interest reader is referred to~\cite{DeleforgeForbesHoraud2015} for an in-depth description and discussion.
Once initialized, at each iteration $i$, the algorithm alternates between two expectation steps, E-Z and E-W, and two maximization steps, M-GMM and M-mapping:

%For the sake of readability, the current iteration superscript $(i)$ is replaced with a tilde, \textit{e.g}, $\thetavect^{(i)} = \widetilde{\thetavect}$.

\begin{itemize}
\item \textbf{E-W-step:}
\label{par:SOLVA_EWstep} Given the current parameter estimates $\thetavect^{(i-1)}$, the posterior probability
$r^{(i)}_{W|Z}$ in \eqref{eq:r_wz} is fully determined by the distributions
$p(\wvect_n|Z_n=k,\tvect_n,\yvect_n;\thetavect^{(i-1)})$ for all $n$ and $k$, which can be
shown to be Gaussian. Their covariance matrices
$\Smat_k^{w(i)}$ and vector means $\muvect_{nk}^{w(i)}$ are given by
\begin{align}
 \label{eq:SOLVA_EWstep3}
 \Smat_k^{w(i)} &= {\left(\Imat+{\Amat_k^{w(i-1)}}\tp {\Sigmamat_k^{(i-1)}}\inverse\Amat_k^{w(i-1)}\right),}\inverse \\
  \muvect_{nk}^{w(i)} &= \Smat_k^{w(i)} {\Amat_k^{w(i-1)}}\tp {\Sigmamat_k^{(i-1)}}\inverse \nonumber \\
  & \times \left(\yvect_n-\Amat_k^{t(i-1)}\tvect_n-\bvect_k^{(i-1)}\right).
\end{align}

\item \textbf{E-Z-step:} \label{par:SOLVA_EZstep} The posterior
probability $r^{(i)}_{Z}$  in \eqref{eq:r_z} is determined by
\begin{align}
 \label{eq:SOLVA_EZstep}
 r^{(i)}_{nk}&=p(Z_n=k|\tvect_n,\yvect_n;\thetavect^{(i-1)}) = \nonumber\\
                   &=\frac{\pi_k^{(i-1)}p(\yvect_n,\tvect_n|Z_n=k;\thetavect^{(i-1)})}{\textstyle\sum_{j=1}^K\pi_j^{(i-1)}p(\yvect_n,\tvect_n|Z_n=j;\thetavect^{(i-1)})}
\end{align}
for all $n$ and $k$, where
\begin{align}
p(\yvect_n, \tvect_n|Z_n&=k;\thetavect^{(i-1)}) \nonumber \\
&=\mathcal{N}(\tvect_n;\cvect_k^{t},\Gammamat_k^{t})
\mathcal{N}(\yvect_n;\dvect_k,\Phivect_k),
\end{align}
with:
\begin{align*}
\dvect_k &= \Amat_k^{t(i-1)}\tvect_n+\bvect_k^{(i-1)},\\
\Phivect_k &= \Amat_k^{w(i-1)}{\Amat_k^{w(i-1)}}\tp + \Sigmamat_k^{(i-1)}.
\end{align*}
The maximization (\ref{eq:Q-PSM-EM}) can then be
performed using the posterior probabilities $r^{(i)}_{nk}$ and
the sufficient statistics $\muvect_{nk}^{w(i)}$
and $\Smat_k^{w(i)}$. We use the following
notations: $\rho_{nk}^{(i)}=r^{(i)}_{nk} / \sum_{n=1}^Nr^{(i)}_{nk}$ and
$\xvect^{(i)}_{nk}=[\tvect_n;\muvect_{nk}^{w(i)}]\in\mathbb{R}^L$.
The M-step can be divided into two separate steps.

\item \textbf{M-GMM-step:}
The updating of parameters $\pi^{(i)}_k,
\cvect_k^{t(i)}$ and
$\Gammamat_k^{t(i)}$ correspond to those of a standard Gaussian mixture model
on $\Tvect_{1:N}$, so that we get straightforwardly:
\begin{align}
\cvect_k^{t(i)} &= \sum_{n=1}^N\rho_{nk}^{(i)} \tvect_n, \\
\Gammamat_k^{t(i)}&= \sum_{n=1}^N \rho_{nk}^{(i)} (\tvect_n-\cvect_k^{t(i)})(\tvect_n-\cvect_k^{t(i)})\tp \\
\pi^{(i)}_k &=\frac{\sum_{n=1}^Nr^{(i)}_{nk}}{N}.
\end{align}

\item \textbf{M-mapping-step:}
The updating of mapping parameters $\{\Amat_k, \bvect_k, \Sigmamat_k\}_{k=1}^K$ is also in closed-form. The affine transformation matrix is updated with:
\begin{equation}
\label{eq:Ak_up}
\Amat^{(i)}_k=\Ymat^{(i)}_k{\Xmat^{(i)}_k}\tp(\Smat_k^{\textrm{x}(i)}+\Xmat^{(i)}_k{\Xmat^{(i)}_k}\tp)^{-1}
\end{equation}
where:
\begin{align}
\label{eq:Xk}
 \Xmat^{(i)}_k & = %\frac{1}{\sqrt{r^{(i)}_k}}
 \left(\sqrt{\rho_{1k}^{(i)}}(\xvect^{(i)}_{1k}-\xvect^{(i)}_k), \dots, \sqrt{\rho_{nk}^{(i)}}(\xvect^{(i)}_{Nk}-\xvect^{(i)}_k)\right), \nonumber\\
 \Ymat^{(i)}_k & = %\frac{1}{\sqrt{r^{(i)}_k}}
 \left(\sqrt{\rho_{1k}^{(i)}}(\yvect_1-\yvect^{(i)}_k),\dots, \sqrt{\rho_{nk}^{(i)}}(\yvect_N-\yvect^{(i)}_k)\right), \nonumber \\
 \xvect^{(i)}_k & = \sum_{n=1}^N \rho_{nk}^{(i)} \xvect^{(i)}_{nk}, \nonumber \\
 \yvect^{(i)}_k &= \sum_{n=1}^N \rho_{nk}^{(i)} \yvect_n, \nonumber \\
 \Smat_k^{\textrm{x}(i)} &=
  \begin{pmatrix}
    \zerovect & \zerovect\\
    \zerovect & \Smat_k^{w(i)}
  \end{pmatrix}.\nonumber
\end{align}
%%%%%%%%%%% FOR ARXIV %%%%%%%%%%%%%
% $
%  \Xmat^{(i)}_k = \frac{1}{r^{(i)}^{1/2}_k}\left[r^{(i)}^{1/2}_{1k}(\xvect^{(i)}_{1k}-\xvect^{(i)}_k) \dots r^{(i)}_{Nk}^{1/2}
%  (\xvect^{(i)}_{Nk}-\xvect^{(i)}_k)\right],
%  \Ymat^{(i)}_k = \frac{1}{r^{(i)}_k^{1/2}}\left[r^{(i)}_{1k}^{1/2}(\yvect_1-\yvect^{(i)}_k)
%  \dots r^{(i)}_{Nk}^{1/2}(\yvect_N-\yvect^{(i)}_k)\right],\\
% $
%%%%%%%%%%%%%%%%%%%%%%%%%%%%%%%%%%%
The intercept parameters are updated with:
\begin{equation}
 \bvect^{(i)}_k =
 \sum_{n=1}^N\rho_{nk}^{(i)} (\yvect_n-\Amat^{(i)}_k\xvect^{(i)}_{nk}).
\end{equation}
The noise covariance matrices are updated with:
\begin{align}
 \label{eq:SOLVA_Sigmak}
 &\Sigmamat^{(i)}_k = \operatorname{diag}\Bigl\{\Amat_k^{w(i)}\Smat_k^{w(i)}{\Amat_k^{w(i)}}\tp+ \\
 &\sum_{n=1}^N\rho_{nk}^{(i)} (\yvect_n-\Amat^{(i)}_k\xvect^{(i)}_{nk}- \bvect^{(i)}_k)
 (\yvect_n-\Amat^{(i)}_k\xvect^{(i)}_{nk}- \bvect^{(i)}_k)\tp\Bigr\}\nonumber
\end{align}
where the $\operatorname{diag}\{\cdot\}$ operator sets all the off-diagonal entries to $0$.
\item \textbf{Initialization:} Initial parameters $\thetavect^{(0)}$ are obtained by fitting a GMM with $K$ components to the joint output-input training dataset $\{\tvect_n,\yvect_n\}_{n=1}^N$.
\end{itemize}
Note that the following derivations are also valid for the estimation of the parameter set $\thetavect$ in (\ref{eq:theta-def}), which corresponds to $L_w=0$, hence the E-W step disappears.

% or


%\appendices
%% you can choose not to have a title for an appendix
%% if you want by leaving the argument blank
%\section{}



\bibliographystyle{IEEEtran}
%\bibliography{IEEEabrv,bibrefs}


\end{document}